%% file: main.tex
\documentclass[runningheads]{llncs}

\usepackage{eccv}

\usepackage{eccvabbrv}

\usepackage{graphicx}
\usepackage{booktabs}

\usepackage[accsupp]{axessibility}  

\usepackage{hyperref}

\usepackage{orcidlink}
\usepackage{bm}
\usepackage{bbm}
\usepackage{algorithm}
\usepackage{algorithmic}
\usepackage{multicol}
\usepackage{multirow}
\usepackage{makecell}
\usepackage{colortbl}

\begin{document}

\title{SCOMatch: Alleviating Overtrusting\\
in Open-set Semi-supervised Learning} 


\author{Zerun Wang\inst{1}\orcidlink{0009-0004-7762-5356} \and
Liuyu Xiang\inst{2}\orcidlink{0000-0001-8486-6255}, Lang Huang\inst{1}\orcidlink{0000-0002-3405-0298}, Jiafeng Mao\inst{1}\orcidlink{0009-0003-0907-7522}, \\
Ling Xiao\inst{1}\orcidlink{0000-0002-4650-8841} \and Toshihiko Yamasaki\inst{1}\orcidlink{0000-0002-1784-2314}}

\authorrunning{Z.~Wang et al.}

\institute{The University of Tokyo \\
\email{\{ze\_wang, langhuang, ling, yamasaki\}@cvm.t.u-tokyo.ac.jp}\\
\email{mao@hal.t.u-tokyo.ac.jp}
 \and
Beijing University of Posts and Telecommunications\\
\email{xiangly@bupt.edu.cn}}

\maketitle

\begin{abstract}
  Open-set semi-supervised learning (OSSL) leverages practical open-set unlabeled data, comprising both in-distribution (ID) samples from seen classes and out-of-distribution (OOD) samples from unseen classes, for semi-supervised learning (SSL). Prior OSSL methods initially learned the decision boundary between ID and OOD with labeled ID data, subsequently employing self-training to refine this boundary. These methods, however, suffer from the tendency to overtrust the labeled ID data: the scarcity of labeled data caused the distribution bias between the labeled samples and the entire ID data, which misleads the decision boundary to overfit. The subsequent self-training process, based on the overfitted result, fails to rectify this problem. In this paper, we address the overtrusting issue by treating OOD samples as an additional class, forming a new SSL process. 
  Specifically, we propose SCOMatch, a novel OSSL method that 1) selects reliable OOD samples as new labeled data with an OOD memory queue and a corresponding update strategy and 2) integrates the new SSL process into the original task through our \textbf{S}imultaneous \textbf{C}lose-set and \textbf{O}pen-set self-training. SCOMatch refines the decision boundary of ID and OOD classes across the entire dataset, thereby leading to improved results. Extensive experimental results show that SCOMatch significantly outperforms the state-of-the-art methods on various benchmarks. The effectiveness is further verified through ablation studies and visualization. Our code will be available at \url{https://github.com/komejisatori/SCOMatch}.
\end{abstract}

\section{Introduction}
\label{sec:intro}

Semi-supervised learning (SSL)~\cite{zhou2021semi} significantly enhances the efficacy of numerous computer vision tasks by leveraging a large number of available unlabeled data for training. A common approach is to train the model on labeled data while using the model to assign pseudo-labels to unlabeled data for self-training.

Conventional SSL methods~\cite{zhang2021flexmatch,wang2022freematch,li2021comatch,zheng2022simmatch,sohn2020fixmatch} assume that the classes in unlabeled data are the same as those in labeled data, a.k.a., in-distribution (ID) classes. This assumption greatly differs from real-world applications, where unlabeled data often contains some unseen classes, \ie, the out-of-distribution (OOD) class. Samples from the OOD class can affect the SSL performance, as the model can only assign ID classes to the samples, leading to the model learning incorrect information.

\begin{figure}[t]
  \centering
  \includegraphics[width=0.99\linewidth]{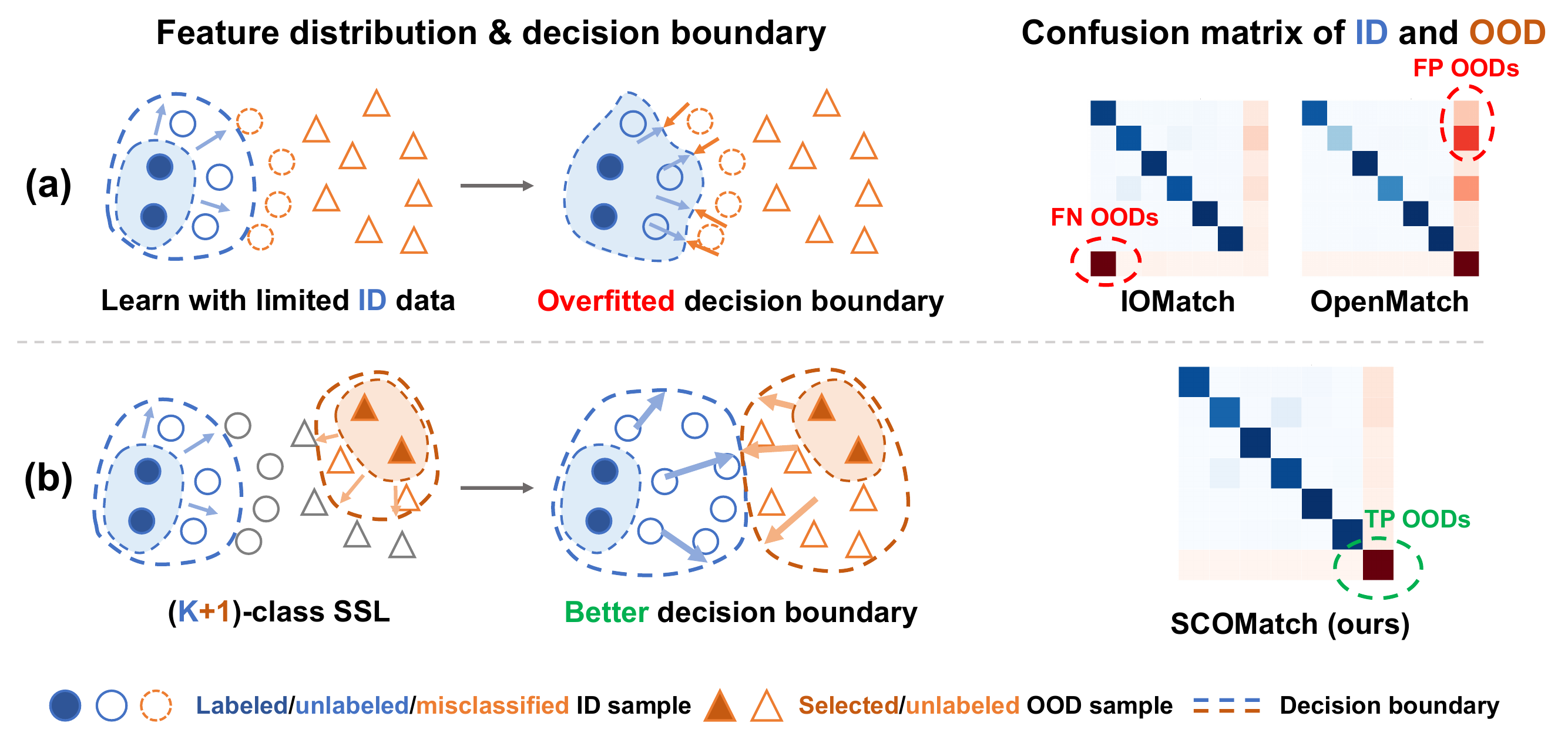}
  \caption{Comparison between prior OSSL methods and SCOMatch on CIFAR-10 with six ID classes. \textbf{(a):} Prior methods overtrust labeled ID data, leading to overfitted decision boundaries that self-training cannot rectify. This results in more false positive or negative OODs in the red circles of the confusion matrix. \textbf{(b):} SCOMatch selects reliable OOD samples for ($K$+1)-class SSL, achieving higher accuracy for both ID and OOD classes.}
  \label{fig:fig0} 
\end{figure}

Open-set semi-supervised learning (OSSL) methods~\cite{guo2020safe,chen2020semi,yu2020multi,huang2021trash,fan2023ssb} are proposed to address this issue. The essence of OSSL lies in identifying OOD samples in unlabeled data so that they can be excluded during the SSL process. Previous OSSL methods typically obtain the ability to identify OOD samples by learning from labeled ID data: they apply OOD detection~\cite{li2023iomatch,saito2021openmatch} or prototype learning~\cite{ma2023rethinking} methods to learn from the distribution of labeled ID data and get the decision boundary that distinguishes ID and OOD samples. Recent OSSL methods~\cite{saito2021openmatch,he2022safe} further apply self-training methods to improve this decision boundary on unlabeled data.

These methods have improved the effectiveness of SSL using open-set unlabeled data. However, they overtrust the labeled ID data. As illustrated in Figure~\ref{fig:fig0}(a): in OSSL, labeled ID data is limited, leading to a distribution bias with the ID data in the entire dataset. As a result, they learn a suboptimal decision boundary overfitted to the labeled data. The following self-training can only be based on this overfitted boundary, thus failing to address this issue. The overtrust problem results in more false-positive or false-negative OODs, as shown in the confusion matrices of Figure~\ref{fig:fig0}(a), thereby compromising the performance.

Meanwhile, we noticed that overtrusting does not exist in conventional SSL tasks. We found that the key difference is that in conventional SSL, all classes have labeled data from the beginning. Consequently, each class can start to refine its decision boundary rather than depending on the boundary of other classes. Inspired by this, we aim to treat `OOD' as a new class with its own labeled data and form a new ($K$+1)-class SSL process ($K$ is the number of ID classes). As illustrated in Figure~\ref{fig:fig0}(b): we first select some high-quality OOD samples as the labeled data of the ($K$+1)-th class. Supervised by this extended labeled dataset, we start a ($K$+1)-class SSL process. Thus, both ID and OOD classes can refine their decision boundary independently across the entire dataset. This results in better discrimination of both ID and OOD as shown in the confusion matrix of Figure~\ref{fig:fig0}(b). While some recent OSSL works~\cite{li2023iomatch,ma2023rethinking} also treat `OOD' as a new class, their OOD samples are still based on the decision boundary learned with limited labeled ID data. Thus, their methods do not deviate from the scope of previous methods that overtrust labeled data while ours can alleviate this issue.

Following our solution of forming a ($K$+1)-class SSL process, there are two challenges. First, how to select reliable OOD samples as the new labeled data, since all the OODs were originally unlabeled in this task. Second, how to integrate the new ($K$+1)-class SSL task into the original $K$-class task, as the latter is the objective of the OSSL task, and the new task should not affect its performance. To address these challenges, we propose SCOMatch, which consists of the following key designs: 
1) for selecting reliable OOD samples, we propose employing an OOD memory queue with a corresponding update strategy to maintain reliable OOD samples as additional labeled data;
2) for integrating the new ($K$+1)-class SSL with the original task, we propose a simultaneous close-set and open-set self-training process, allowing the open-set and close-set self-training to be jointly optimized on a single classification head. 

The experimental results on various benchmarks show that SCOMatch significantly outperforms prior methods and achieves state-of-the-art performance. We also prove the effectiveness of our proposed components through ablation studies and analysis.

Overall, our contributions can be summarized as follows:
\begin{itemize}
    \item We identify the issue of overtrusting limited labeled ID data in prior OSSL methods and alleviate it by treating OOD as a new class and forming a new ($K$+1)-class SSL process. 
    \item We propose a novel OSSL method, SCOMatch, which tackles the challenges of reliable OOD sample selection and integrates the new ($K$+1)-class SSL with the original task through the designed components. 
    \item SCOMatch significantly outperforms prior OSSL methods on various OSSL benchmarks. In particular, we improve the close-set accuracy by 13.4\% on TinyImageNet. We also conduct comprehensive experiments and visualization to prove the effectiveness of each component.
    
\end{itemize}

\section{Related Work}

\subsection{Semi-supervised learning}
Semi-supervised learning (SSL) has attracted considerable attention in recent years. The mainstream SSL methods typically train the model on labeled data while employing consistency regularization~\cite{bachman2014learning} to utilize unlabeled data. FixMatch~\cite{sohn2020fixmatch} is one of the influential SSL methods. It summarizes some early works~\cite{berthelot2019mixmatch,xie2020unsupervised,rasmus2015semi,berthelot2019remixmatch} and provides a simple and efficient framework. Some of the following SSL methods contribute to various parts of this framework: FlexMatch~\cite{zhang2021flexmatch} and FreeMatch~\cite{wang2022freematch} dynamically adjust each class's threshold during different training stages based on learning performance. CoMatch~\cite{li2021comatch} and SimMatch~\cite{zheng2022simmatch} apply contrastive learning strategies to improve semantic similarity in the same class, thus improving the pseudo-label quality. Some works also focus on addressing issues such as class-imbalance~\cite{yu2022inpl,guo2022class} and domain adaptation~\cite{berthelot2021adamatch,chen2015deep}. However, all these SSL methods are based on the assumption that no OOD samples exist in unlabeled data. Therefore, the OOD samples in real-world unlabeled data will affect their performance.


\subsection{Open-set semi-supervised learning}
Open-set semi-supervised learning, also called safe semi-supervised learning, aims to address the problem of OOD samples in unlabeled data. The key point is to identify OOD samples. Early works attempt to remove OOD samples or mitigate their effect on the SSL process. DS3L~\cite{guo2020safe} assigns a learnable weight on unlabeled data, thereby reducing the contribution of OOD samples to the training loss. UASD~\cite{chen2020semi} applies a threshold on the classification confidence score to identify and remove OOD samples during SSL. MTCF~\cite{yu2020multi} applies a binary classification head to distinguish ID and OOD samples. T2T~\cite{huang2021trash} uses a cross-modal matching branch for filtering out OOD samples. Later works discover that the OOD samples in unlabeled data, once identified, can also be used for training. OpenMatch~\cite{saito2021openmatch} applies the one-vs-all (OVA) classification head~\cite{saito2021ovanet} to detect OOD samples. The detected samples will used for SSL via consistency regularization. SAFE-STUDENT~\cite{he2022safe} uses energy discrepancy for labeling OOD samples. Then, the model will be trained to increase the uncertainty of these samples. These methods have improved the OSSL performance. However, as the OOD classification modules are trained on labeled data first, they face the overtrusting issue, as we pointed out. 

Although several recent works treat `OOD' as a new class, they do not deviate from the scope of past methods. IOMatch~\cite{li2023iomatch} calculates the open-set probability score using the OVA head's output and then optimizes an ($K$+1)-class classification head. However, the OVA head still relies on labeled ID data for training as the same in OpenMatch. Ma~\etal~\cite{ma2023rethinking} apply the prototype learning method to solve the OSSL problem. They use the labeled data to get prototypes for each ID class first and use the fixed prototypes for labeling ID and OOD samples and train a ($K$+1)-way classification head. It is also obvious that the prototypes are based on labeled data only. In contrast, we don't trust the result trained on labeled ID data only. Instead, we start a new ($K$+1)-class SSL process with labeled IDs and selected OODs simultaneously from the beginning. 

OSSL methods also take some techniques from the OOD detection task~\cite{hendrycks2016baseline}. However, OOD detection aims to train on a large number of labeled ID data to distinguish OOD samples, which is different from the OSSL setting.

\section{Methodology}

We aim to solve the overtrusting problem by treating OOD as a new class with its own labeled data and forming a ($K$+1)-class SSL task. To achieve this, we propose to 1) select reliable OOD samples as the new labeled data and 2) integrate the new ($K$+1)-class SSL into the original $K$-class task. In this section, we first introduce the preliminary of SSL, then the overall framework of SCOMatch, and finally, we separately explain how we select reliable OOD samples and combine the two tasks.

\subsection{Preliminary}
OSSL aims to solve the open-set problem within the SSL frameworks. In this setting, we consider the models trained on both labeled dataset  $\mathcal{D}^x = \{(\bm{x}_i, y_i)\}_{i=1}^{N}$ and unlabeled dataset $\mathcal{D}^u = \{(\bm{u}_i)\}_{i=1}^{M}$. Here, $\bm{x}_i$ and $\bm{u}_i$ represent the images, $y_i\in\{1,2,...,K\}$ denotes the label within $K$ ID classes, and $N\ll M$ since the labeled data is limited. 
Similar to most recent OSSL works, we utilize some basic SSL techniques from FixMatch~\cite{sohn2020fixmatch}. Suppose that we have a backbone $\mathcal{F}$ and a classification head $\mathcal{C}$ parametrized by weights $\mathbf{C}$ and subsequently normalized by a softmax function. In each training iteration, we have a batch of $B$ labeled samples $\mathcal{X}=\{(\bm{x_b},y_b)\}_{b=1}^{B}$ for supervised training, and a batch of $\mu B$ unlabeled samples $\mathcal{U}=\{(\bm{u_b})\}_{b=1}^{\mu B}$ for self-training. Here, $\mu$ controls the relative size of $\mathcal{X}$ and $\mathcal{U}$.

For the supervised loss on labeled data $\mathcal{L}_{s}$, we apply a weak augmentation on images and use the standard cross-entropy $H$ to optimize the model:
\begin{equation}
\label{eq:eq1}
\mathcal{L}_{s} = \frac{1}{B}\sum_{i=1}^{B}H(y_i, \mathcal{C}(\mathcal{F}(\bm{x}_i^{w}))),
\end{equation}
where $\bm{x}_i^{w}$ denotes the weakly-augmented images.

For the self-training loss on unlabeled data $\mathcal{L}_{u}$, we first get the model’s prediction given an unlabeled sample with weak augmentation: $\bm{q}_i = \mathcal{C}(\mathcal{F}(\bm{u}_i^w))$. Then, we apply thresholding to select those samples with high confidence, instantiated by an indicator function $\mathbbm{1}(\max(\bm{q}_i)>\tau)$ where $\tau$ is a manual threshold. The selected predictions are then converted to one-hot pseudo-labels $\hat{q}_i = \arg \max(\bm{q}_i)$. For training, we apply strong augmentation on the same samples and use cross-entropy for consistency regularization:
\begin{equation}
\label{eq:eq2}
\mathcal{L}_{u} = \frac{1}{\mu B}\sum_{i=1}^{\mu B}\mathbbm{1}(\max(\bm{q}_i)>\tau)H(\hat{q}_i, \mathcal{C}(\mathcal{F}(\bm{u}_i^s))),
\end{equation}
where $\bm{u}_i^s$ denotes the unlabeled image with strong augmentation.

The model is optimized by the total combined loss of $\mathcal{L}_{s}+\lambda \mathcal{L}_{u}$, where $\lambda$ is the hyperparameter controlling the trade-off between these two terms. Following prior works, the Exponential Moving Average (EMA)~\cite{tarvainen2017mean} technique is applied to enhance training stability.

\subsection{Overall framework}

\begin{figure*}[t]
  \centering
  \includegraphics[width=1\linewidth]{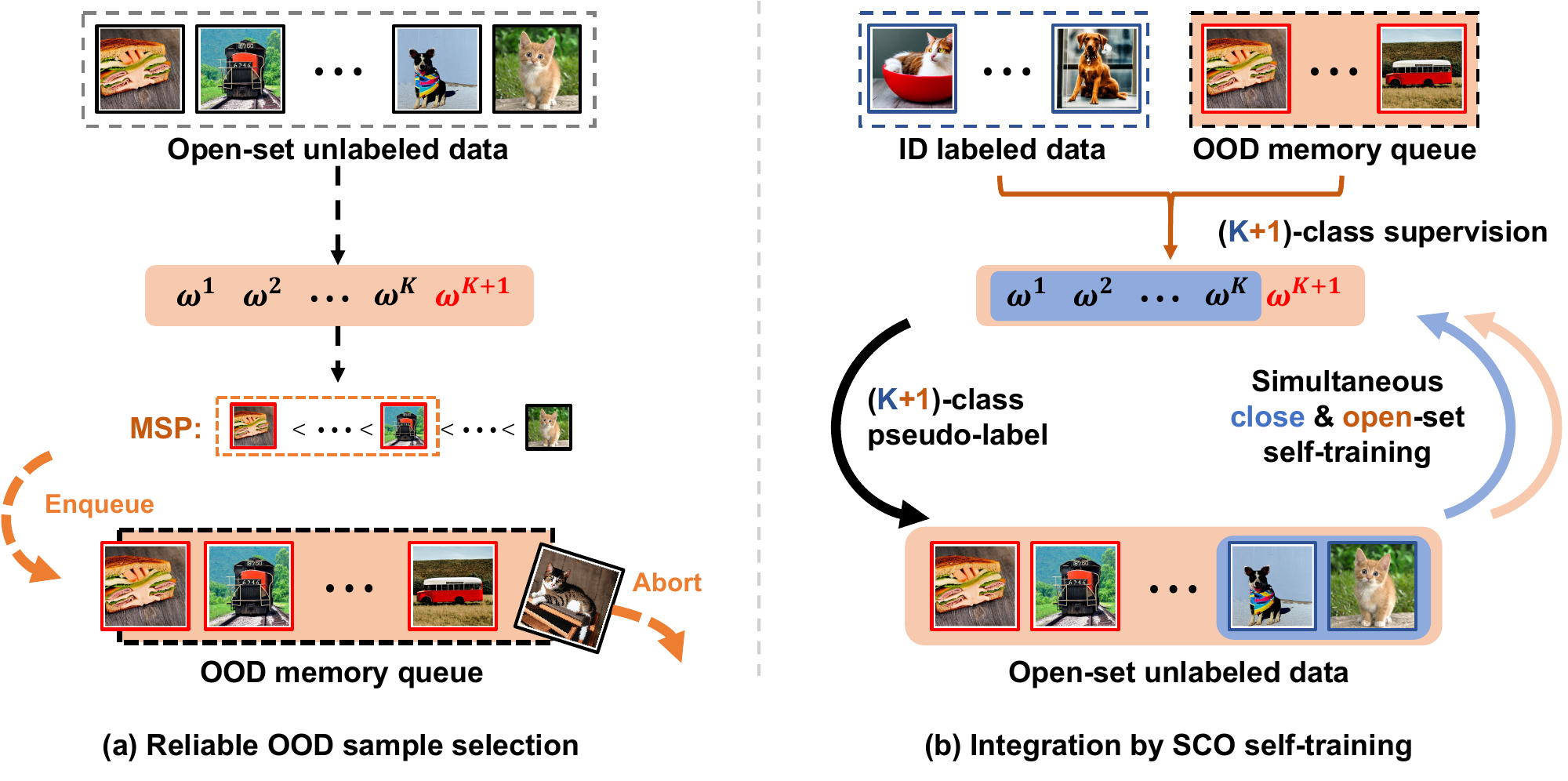}
  \caption{The training process of SCOMatch. The ($K$+1)-classification head is the only head structure in our model (the backbone is not depicted for simplification). \textbf{(a)} The OOD sample selection by our proposed OOD memory queue and corresponding update strategy. \textbf{(b)} The integration of the original $K$-class task and the new ($K$+1)-class SSL by our simultaneous close-set and open-set self-training. These two processes function concurrently with the same model, but we separate them for the clarity of explanation. Here, the animal classes are ID and others are OOD.}
  \label{fig:fig2}
\end{figure*}


We propose SCOMatch with 1) the OOD memory queue and corresponding update strategy and 2) simultaneous close-set and open-set self-training (SCO self-training) to alleviate overtrusting. Figure~\ref{fig:fig2} presents the structure and training process of our SCOMatch. The ($K$+1)-class classification head $\mathcal{C}$ with parameters $\mathbf{C}=[\bm{w}^{1}, \bm{w}^{2}, ..., \bm{w}^{K+1}]$ is the only head structure in SCOMatch, where first $K$ elements are for the $K$ ID classes and the last dimension $\bm{w}^{K+1}$ is newly introduced for the `OOD' class. Figure~\ref{fig:fig2}(a) illustrates the OOD memory queue and corresponding date strategy for reliable OOD sample selection. Figure~\ref{fig:fig2}(b) illustrates the integration of original $K$-class tasks and the new ($K$+1)-class SSL task by our SCO self-training. We provide detailed introductions to each part in the subsequent sub-sections.

\subsection{Reliable OOD sample selection}
The cornerstone of our proposed method is to construct a new ($K$+1)-class dataset. This necessitates the selection of reliable, high-quality OOD samples. However, there are two major problems: 1) All the OOD samples are unlabeled, inevitably introducing noise during the selection; 2) Overtrusting, as in prior OSSL methods, should be avoided when we select from the unlabeled data. 
To alleviate these issues, we propose the following well-designed OOD memory queue and corresponding update strategy to obtain reliable OOD samples, as illustrated in Figure~\ref{fig:fig2}(a).

\noindent\textbf{The OOD memory queue.} We construct a first-in first-out memory queue  $\mathcal{D}^{m}=\{(\bm{o}_i, y_i^{ood})\}_{i=1}^{N_{m}}$ with a relatively small size $N_{m}$ to store the selected images $\{\bm{o}_i\}$ that belong to the OOD class with $y^{ood} = K+1$.

\noindent\textbf{Update strategy.} 
Prior works use the OOD detection head or scoring methods, which are learned with limited ID labeled data, to detect OODs. In contrast, we seek to leverage the knowledge of the entire ($K$+1)-class task to avoid overtrusting. Considering Maximum Softmax Probability (MSP)~\cite{hendrycks2016baseline} is a reliable scoring method for detecting OODs, we adapt it to our ($K$+1)-class head $\mathcal{C}$ to select OODs from unlabeled data:
\begin{equation}
    {\rm MSP}^{\mathcal{C}}(\bm{u}) = \max_{i\in\{1,..,K\}} \mathcal{C}(\mathcal{F}(\bm{u^w}))^i,
\label{eq:eq3}
\end{equation}
which means that we consider the maximum probability among the first $K$ classes of the entire ($K$+1)-class head. 

During training, we sort the MSP of samples in each unlabeled batch and push $K_{m}$ samples with the smallest MSP into the OOD memory queue $D_{m}$, since samples with smaller MSP are more likely to be OOD. Meanwhile, the oldest sample in the queue will be aborted since it is first-in first-out.

Our proposed OOD memory queue and update strategy can tackle the above challenges: 1) The first-in first-out scheme with relatively small $N_{m}$ avoids the noisy or overtrusted sample remaining in the queue for a long time. 2) We keep $K_{m}$ relatively small. Therefore, only the samples with the lowest MSP score (indicating high confidence of being OOD) will be selected. Experimental results prove our design's effectiveness and the selected samples' quality.

\subsection{Integration by SCO self-training}

With the selected reliable OOD samples, we construct a new ($K$+1)-class labeled data for OSSL. 
However, the $K$-class close-set and ($K$+1)-class open-set tasks inherently conflict as they attempt to cluster the embedded features from the backbone into different distributions. We empirically found that the two tasks will interfere with each other if we follow the usual multi-task paradigm and use two separate heads for each task.

This is solved by our proposed SCO self-training as illustrated in Figure~\ref{fig:fig2}(b): with the new labeled data, the model is only supervised by the ($K$+1)-class task. Then, the model is used for generating ($K$+1)-class pseudo-label. After that, we simultaneously use the first $K$-class pseudo-labels for close-set self-training on $\mathcal{C}^K$ and all the ($K$+1)-class pseudo-labels for open-set self-training on $\mathcal{C}$.  Our SCO self-training allows two classification tasks with conflicting class numbers to be trained together and jointly serve the ultimate goal of the OSSL task.

It is worth noting that while prior OSSL methods are supervised by $K$-class labeled ID data, our head $\mathcal{C}$ is only supervised by the new ($K$+1)-class labeled data combination, which avoids the head overfitting to the labeled ID data. Meanwhile, although OOD samples actually originate from multiple categories, we find that treating OODs as a single class can address the overtrusting issue and enhance the effectiveness of OSSL. This is because the goal of OSSL tasks is not distinctions within OOD categories, and having the model pay additional attention to classifying among OODs within a limited parameter space could potentially affect the distinction between ID classes.

\noindent\textbf{(K+1)-class supervision.} For each iteration, we use one batch of labeled ID data $\bm{x}\in\mathcal{X}$ and randomly select the same number of OOD samples from the OOD memory queue $\bm{o}\in{D^m}$ for ($K$+1)-class supervision on the entire head $\mathcal{C}$:
\begin{equation}
\label{eq:eq4}
\mathcal{L}^{id}_{s} = \frac{1}{B}\sum_{i=1}^{B}H(y_i, \mathcal{C}(\mathcal{F}(\bm{x}_i^w))),
\end{equation}
\begin{equation}
\label{eq:eq5}
\mathcal{L}^{ood}_{s} = \frac{1}{B}\sum_{i=1}^{B}H(y^{ood}_i, \mathcal{C}(\mathcal{F}(\bm{o}_i))).
\end{equation}
The two losses, $\mathcal{L}^{id}_{s}$ for using labeled ID data and $\mathcal{L}^{ood}_{s}$ for using selected OOD samples, are combined together as the supervision loss $\mathcal{L}_{s}$ in SSL to train the entire head. We use the cross-entropy loss following Eq.~\ref{eq:eq1}. 


\noindent\textbf{(K+1)-class pseudo-labeling.} With the supervision of our ($K$+1)-class labeled data, we use $\mathcal{C}$ for ($K$+1)-class pseudo-labeling with the thresholding method $\mathbbm{1}(\max(\bm{q}_i)>\tau')$ defined in the preliminary. Here, $\tau'=\{\tau, \tau^{ood}\}$ denotes that we use $\tau^{ood}$ of the new OOD class and the original SSL threshold $\tau$ for ID classes. Following other OSSL methods, we adjust the threshold $\tau^{ood}$ dynamically since the OOD class may actually contain more than one unseen class, its learning difficulty is different from ID classes. Inspired by the Curriculum Pseudo Labeling (CPL)~\cite{zhang2021flexmatch}, we adjust the threshold $\tau^{ood}$ by:
\begin{equation}
\label{eq:eq6}
    \tau_{t}^{ood}=\frac{\sigma_t(c^{ood})}{\sum_{i=1}^{K} \sigma_{t}(c^{i})}\cdot \tau,
\end{equation}
where $\sigma_t(c^{ood})$ denotes the number of OOD samples with the confidence score above the original threshold $\tau$. $\sum_{i=1}^{K} \sigma_{t}(c^{i})$ denotes the number of samples above $\tau$ across all the ID classes. At training iteration $t$, the threshold of the OOD class $\tau_{t}^{ood}$ is scaled by the ratio between $\sigma_t(c^{ood})$ and $\sum_{i=1}^{K} \sigma_{t}(c^{i})$. We set upper and lower bounds $[\tau_{min},\tau]$ for $\tau^{ood}$ to avoid unstable training.

\noindent\textbf{Open-set self-training.}
We directly use the ($K$+1)-class pseudo label on $\mathcal{C}$ for open-set self-training. We use both weak-augmented and strong-augmented images $\bm{u}_i^{w},\bm{u}_i^{s}$:
\begin{equation}
\label{eq:eq7}
\mathcal{L}^{open}_{u} = \frac{1}{2\mu B}\sum_{i=1}^{\mu B}\mathbbm{1}(\max(\bm{q}_i)>\tau')\sum_{\bm{u}_i \in \{\bm{u}_i^w,\bm{u}_i^s\}}H(\hat{q}_i, \mathcal{C}(\mathcal{F}(\bm{u}_i))),
\end{equation}
where $\bm{q}_i = \mathcal{C}(\mathcal{F}(\mathbf{u}_i^{w}))$ and $\hat{q}_i = \arg \max(\bm{q}_i)$ denotes the pseudo-label process defined in the preliminary.

\noindent\textbf{Close-set self-training.} 
As the ultimate goal of OSSL is to better classify ID samples, we find conducting additional close-set self-training on $\mathcal{C}^K$, which is the first $K$ dimensions of $\mathcal{C}$, rather than another individual head can further enhance performance. This may be because the ID samples further correct the distribution of $\mathcal{C}^K$, and improve the accuracy of selecting OOD samples for the OOD memory queue based on MSP. For the close-set self-training loss $\mathcal{L}^{close}_{u}$, we filter the OODs in pseudo label:
\begin{equation}
\label{eq:eq8}
\mathcal{L}^{close}_{u} = \frac{1}{\mu B}\sum_{i=1}^{\mu B}\mathbbm{1}(\hat{q}_i\leq K)\mathbbm{1}(\max(\bm{q}_i)>\tau')H(\hat{q}_i, \mathcal{C}^{K}(\mathcal{F}(\bm{u}_i^{s}))),
\end{equation}
where $\mathbbm{1}(\hat{q}_i\leq K)$ denotes that we filter out the OODs in pseudo-label.

\begin{algorithm*}[t]
\caption{Training process of our SCOMatch}
\label{alg:alg1}
\textbf{Input}: Labeled data: $\mathcal{D}^x = \{(\bm{x}_i, y_i)\}_{i=1}^{N}$, Unlabeled data: $\mathcal{D}^u = \{(\bm{u}_i)\}_{i=1}^{M}$\\
\textbf{Parameters}: SSL: $\tau$, $\lambda$, $\alpha$; SCOMatch: number of enqueue samples $K_m$ \\
\textbf{Output}: {Teacher and student model $\theta_t,\theta_s=\{\mathcal{F}, \mathcal{C}\}$}.
\begin{algorithmic}[1] 
\FOR {semi-supervised learning iterations}
\STATE Update the smallest $K_m$ unlabeled samples into the OOD memory queue based on MSP in Eq.~\ref{eq:eq3}. \\
\STATE Compute supervision loss $\mathcal{L}^{id}_{s}$ with labeled ID data by Eq.~\ref{eq:eq4}. \\
\STATE Get OOD samples from OOD memory queue: $\mathcal{D}^{m}=\{(\mathbf{o}_i, y_i^{ood})\}_{i=1}^{N_{m}}$. \\
\STATE Compute supervision loss $\mathcal{L}^{ood}_{s}$ with selected OOD data by Eq.~\ref{eq:eq5}. \\
\STATE Compute open-set self-training loss $\mathcal{L}^{open}_{u}$ by Eq.~\ref{eq:eq7}. \\
\STATE Compute close-set self-training loss $\mathcal{L}^{close}_{u}$ by Eq.~\ref{eq:eq8}. \\
\STATE Jointly training student model with the loss $\mathcal{L}$ by Eq.~\ref{eq:eq9}.\\
\STATE Update the threshold for the OOD class $\tau^{ood}$ using CPL in Eq.~\ref{eq:eq6}.
\STATE Update student model: $\theta_s \gets \theta_s - \nabla \mathcal{L}$.
\STATE Update teacher model by EMA: $\theta_t \gets \alpha \theta_t + (1-\alpha)\theta_s$\;
\ENDFOR
\STATE \textbf{return} Parameters $\theta_t,\theta_s$ of teacher and student model
\end{algorithmic}
\end{algorithm*}

Finally, the model is trained with the losses mentioned above:
\begin{equation}
\label{eq:eq9}
\mathcal{L} = \underbrace{\mathcal{L}^{id}_{s} + \mathcal{L}^{ood}_{s}}_{\rm (K+1)-class\ supervision} + \lambda\underbrace{(\mathcal{L}^{open}_{u} + \mathcal{L}^{close}_{u})}_{\rm SCO\ self-training},
\end{equation}
where $\lambda$ controls the weight of self-training losses. 
Algorithm~\ref{alg:alg1} summarizes the whole training process.

\section{Experiments}

\subsection{Datasets and metrics}
\textbf{Datasets.}
We construct OSSL benchmarks following Ma~\etal~\cite{ma2023rethinking} and OpenMatch~\cite{saito2021openmatch} with five datasets:  MNIST~\cite{lecun1998mnist}, CIFAR-10~\cite{krizhevsky2009learning}, CIFAR-100~\cite{krizhevsky2009learning}, TinyImageNet~\cite{deng2009imagenet}, and IN-30 (a subset of ImageNet~\cite{deng2009imagenet} containing 30 classes). 

Following Ma~\etal~\cite{ma2023rethinking}, we fix the number of labeled and unlabeled images and select various mismatch ratios, which means the ratio of OOD samples in the unlabeled data. For example, a mismatch ratio of 0.3 means 30\% of the unlabeled data are OODs. In detail, we choose six ID classes (`0'--`5') with ten labeled samples per class and 30,000 unlabeled samples for MNIST; six ID classes (the animal classes) with 400 labeled samples per class and 20,000 unlabeled samples for CIFAR-10; the first 50 classes as ID with 100 labeled samples per class and 40000 unlabeled samples for CIFAR-100; the first 100 classes as ID with 100 samples per class and 40,000 unlabeled samples for TinyImageNet.

Following OpenMatch~\cite{saito2021openmatch}, we use all images in each dataset and select various numbers of labeled images and ID classes. For CIFAR-10, we choose the same six ID classes as above. For CIFAR-100, we choose the first 55/80 classes as ID (the classes are sorted by their super-classes first). For IN-30, we choose the first 20 classes as ID. 

\noindent\textbf{Evaluation metrics.}
Following previous OSSL methods, we use close-set classification accuracy, open-set classification accuracy, and Area under the ROC Curve (AUC)~\cite{hendrycks2016baseline} as evaluation metrics. We use the result of $\mathcal{C}^K$ for close-set classification accuracy and $\mathcal{C}$ for the other two metrics. Close-set classification accuracy considers only the accuracy of ID classes in test data. The open-set classification accuracy and AUC additionally measure the ability to identify OOD data. Open-set classification accuracy considers the accuracy of both ID and OOD classes. AUC is the standard evaluation protocol of novelty detection. We use the predicted probability of the ($K$+1)-th class as the score when calculating AUC. We report the mean result with standard deviation over three runs of different random seeds.

\subsection{Implementation details}
For a fair comparison, we use the same backbone network with prior OSSL works for each task. For MNIST, we use the two-layer CNN in DS3L~\cite{guo2020safe}. For CIFAR-10, CIFAR-100, and TinyImageNet, we use the WideResNet28-2~\cite{zagoruyko2016wide}. For IN-30, we use the ResNet-18. 
For hyperparameters, we set a relatively small size $N_m=8 \times K$ for the OOD memory queue regarding the class number $K$. We set $K_m=1$ as the enqueue number per iteration. The SSL hyperparameters settings follow FixMatch. The other hyperparameters and detailed training settings are reported in the supplementary material. 



\subsection{Comparision with SOTA methods}

\input{tabs/tab1}
\input{tabs/tab2}

\input{tabs/tab3}
\input{tabs/tab4}

\noindent\textbf{Various mismatch ratios.} We first
prove the effectiveness of SCOMatch by comparing it with state-of-the-art SSL and OSSL methods with different mismatch ratios. Here, we use $\mu=6$ to align the unlabeled batch size of our rivals\cite{huang2021trash}. Following Ma~\etal~\cite{ma2023rethinking}, we report the close-set classification accuracy of ID classes in Table~\ref{tab:tab1}. SCOMatch outperforms rivals on all the datasets and mismatch ratios, especially on the more challenging CIFAR-100 and TinyImageNet. On the hardest dataset, TinyImageNet, we achieve 54.2\% accuracy under the mismatch ratio of 0.3, which surpasses about 13.4\% on the strongest rival~\cite{ma2023rethinking}. The AUC results in Table~\ref{tab:tab2} evaluate the ability of OOD identification on MNIST with different ratios. 

\noindent\textbf{Various numbers of labeled data and ID classes.} We further prove the effectiveness of SCOMatch with different numbers of labeled data and ID classes. Here, we use $\mu=4$ to align the unlabeled batch size with SSB\cite{fan2023ssb}. Table~\ref{tab:tab3} and Table~\ref{tab:tab4} show the results of close-set classification accuracy and AUC, respectively. The results suggest that SCOMatch still outperforms its rivals.

\subsection{Ablation studies and discussions}

\input{tabs/tab6}


\textbf{Ablation study on the head structure.} We conduct ablation studies to prove the effectiveness of using a single classification head: without `Single head' in Table~\ref{tab:tab6} (Column 2) means that we use two separate classification heads: one for $K$-class, supervised with labeled ID data and performing close-set self-training, the other for ($K$+1)-class, supervised with both labeled ID data and OOD data from our queue, then perform open-set self-training. The results prove that two separate heads actually impact the performance. As we have discussed, the two heads conflict with each other as they attempt to cluster the embedded feature into different distributions. Moreover, only labeled ID data supervises the $K$-class head, causing SCOMatch to overtrust the labeled data.

\noindent\textbf{Ablation study on SCO self-training.} We also conduct ablation studies on each loss component in SCO self-training in Table~\ref{tab:tab6}. The results show that both close-set and open-set self-training contribute to the performance of SCOMatch (Columns 3 and 4). As we have discussed, open-set self-training is necessary for constructing the ($K$+1)-class SSL task, thus improving the performance of identifying OODs with unlabeled data. Meanwhile, close-set self-training can calibrate the distribution of $\mathcal{C}^{K}$ and improve the quality of OOD sample selection. Thus serving the ultimate goal of the OSSL task, \ie, better classifying ID samples. Note that we cannot use the OOD memory queue without $\mathcal{L}^{ood}_{s}$ and SCOMatch will fall back to FixMatch (Column 1). 

\begin{figure*}[t]
\centering
\includegraphics[width=1.0\linewidth]{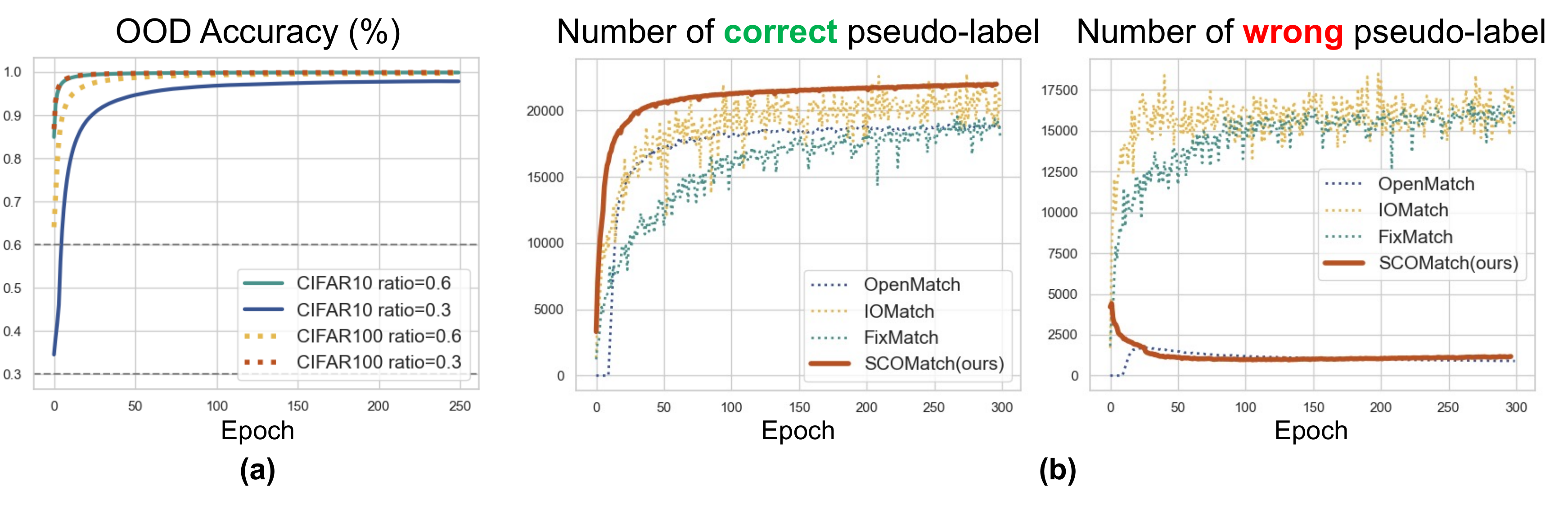}
\caption{\textbf{(a):} The quality of OOD samples in the OOD memory queue during training. The grey dashed line represents the actual ratio of OOD samples in unlabeled data. \textbf{(b):} Correct and wrong pseudo-label number of four methods during training on CIFAR-10 with 50 labeled images.}
\label{fig:fig3}
\end{figure*}

\begin{figure*}[t]
\centering
\subfloat[OpenMatch]{\includegraphics[width=.25\linewidth]{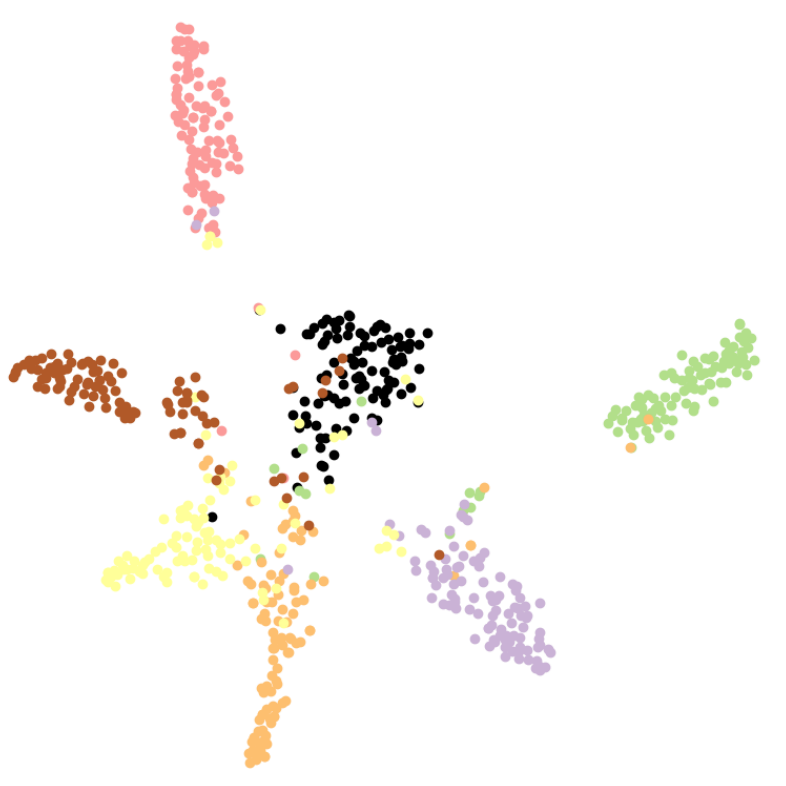}}\hspace{30pt}
\subfloat[IOMatch]{\includegraphics[width=.25\linewidth]{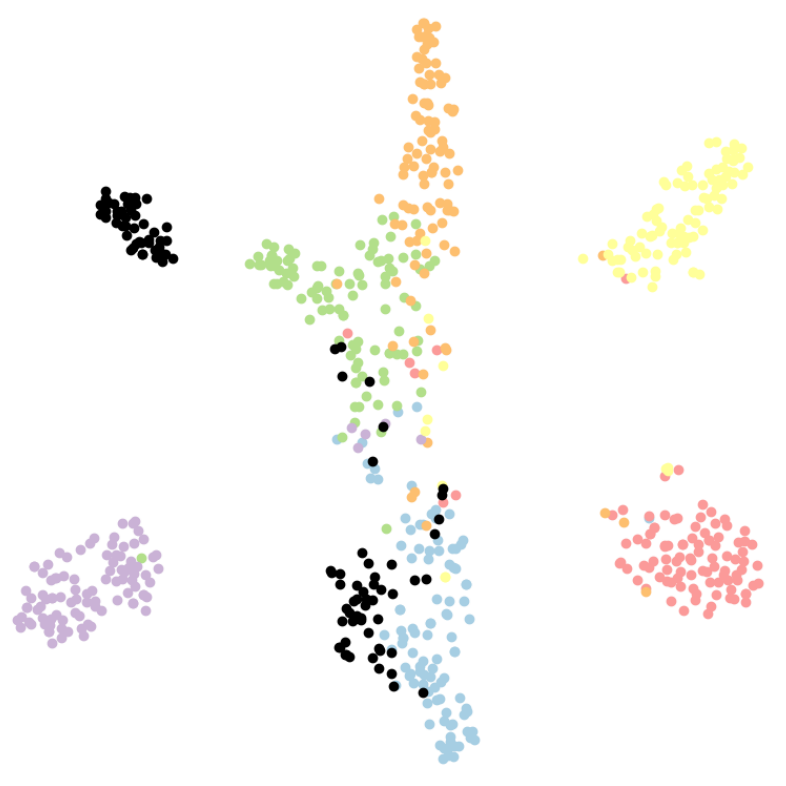}}\hspace{30pt}
\subfloat[SCOMatch (ours)]{\includegraphics[width=.25\linewidth]{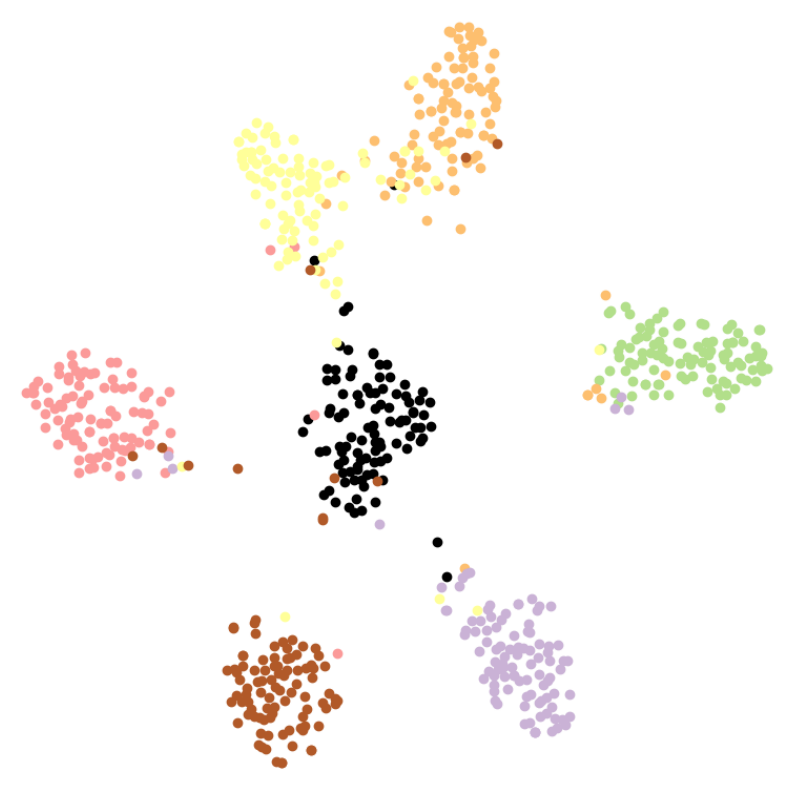}}
\caption{t-SNE visualization results of randomly selected 100 samples from CIFAR-10 test data. Black dots denote the features of OOD samples. Other colors are ID samples.}
\label{fig:fig4}
\end{figure*}

\noindent\textbf{Analysis of the OOD memory queue.} We track the quality of the OOD samples in our queue during training. Figure~\ref{fig:fig3}(a) illustrates the average ratio of real OOD samples in the queue per epoch. The model is trained on CIFAR-10 and CIFAR100 with different mismatch ratios. It can be observed that the queue is noisy in the early stage, but the quality is rapidly improved with training. Note that we actually have 50 OOD classes in CIFAR-100, and the quality remains reliable. This verifies the effectiveness of the proposed OOD memory queue and corresponding update strategy. 

 \noindent\textbf{Can SCOMatch alleviate overtrusting?}
 We further prove that SCOMatch can alleviate the overtrusting problem by exploring the pseudo-label quality. Figure~\ref{fig:fig3}(b) shows the number of correct and wrong pseudo-labels during SSL. Pseudo-labels are the samples from unlabeled data with confidence scores above the SSL threshold. The wrong pseudo-label means the class prediction differs from the annotation, while the correct pseudo-label means the prediction matches the annotation. The results show that SCOMatch generates more correct and less wrong pseudo-labels. Such high-quality pseudo-label leads to better performance. Meanwhile, FixMatch, OpenMatch, and IOMatch either generate a large number of wrong pseudo-labels or fewer correct ones. This phenomenon is consistent with the decision boundary and confusion matrix illustrated in Figure~\ref{fig:fig0}.

\noindent\textbf{Analysis of the feature distribution.}
We apply t-SNE~\cite{van2008visualizing} visualization on the feature after the backbone for randomly selected 100 test samples in CIFAR-10. Figure~\ref{fig:fig4} shows the result of OpenMatch, IOMatch, and SCOMatch. It shows that SCOMatch can form a better decision boundary between different classes, including the OOD class (black dots). The model is trained on CIFAR-10 with 50 labeled ID images and six ID classes as in Table~\ref{tab:tab3}.


\noindent\textbf{Limitations and future direction.} 
SCOMatch only considers the OOD samples of different class spaces within the same domain. The other open-set problems caused by domain difference (\eg, OODs from cartoon drawings or synthetic images) are not considered and will be a potential direction.

\section{Conclusion}

In this paper, we identified the issue of overtrusting limited labeled ID data in prior OSSL methods. To alleviate this, we treat OOD as a new class with its own labeled data and form a new SSL process. Specifically, we proposed SCOMatch, a novel OSSL method that 1)~selects reliable OOD samples as additional labeled data by our OOD memory queue and corresponding update strategy and 2)~integrates the new SSL process with the original task with the proposed simultaneous close-set and open-set self-training. Experimental results on various OSSL benchmarks demonstrated that SCOMatch achieves better performance compared with state-of-the-art methods. We also conducted ablation studies to validate the effectiveness of each component of our method. With SCOMatch, more existing unlabeled data can be effectively leveraged to improve the model's performance without the need for additional manual labeling and filtering. 

\section*{Acknowledgements}
Liuyu Xiang is supported by the National Natural Science Foundation of China (No. 62301066).

%
%
\bibliographystyle{splncs04}
\bibliography{main}
\end{document}

%% file: tabs/tab1.tex
\begin{table*}[t]
\centering
\begin{center}
\caption{Close-set classification accuracy (\%) of different methods on 4 datasets. }
\label{tab:tab1}
\resizebox{\linewidth}{!}{
\begin{tabular}{l|cc|cc|cc|cc|cc}
   \toprule
   \multicolumn{3}{c|}{Dataset}    & \multicolumn{2}{c|}{MNIST}    & \multicolumn{2}{c|}{CIFAR-10} & \multicolumn{2}{c|}{CIFAR-100} & \multicolumn{2}{c}{TinyImageNet} \\
   \midrule
   \multicolumn{3}{c|}{Mismatch ratio}     & 0.3           &  0.6         &  0.3        &  0.6           &    0.3       &   0.6         &   0.3    &            0.6         \\
   \midrule
\multirow{4}{*}{\rotatebox{90}{SSL}} & PL~\cite{lee2013pseudo} & ICML'13 &90.0\scalebox{0.8}{$\pm$0.7} & 86.0\scalebox{0.8}{$\pm$0.6} & 75.8\scalebox{0.8}{$\pm$0.8} & 74.6\scalebox{0.8}{$\pm$0.7} & 60.2\scalebox{0.8}{$\pm$0.3} & 57.5\scalebox{0.8}{$\pm$0.6} & 36.6\scalebox{0.8}{$\pm$0.6} & 35.8\scalebox{0.8}{$\pm$0.4} \\
   & Pi-Model~\cite{sajjadi2016regularization} & NeurIPS'16 & 92.4\scalebox{0.8}{$\pm$0.6} & 86.6\scalebox{0.8}{$\pm$0.5} & 75.7\scalebox{0.8}{$\pm$0.7} & 74.5\scalebox{0.8}{$\pm$1.0} & 59.4\scalebox{0.8}{$\pm$0.3} & 57.9\scalebox{0.8}{$\pm$0.3} & 36.9\scalebox{0.8}{$\pm$0.4} & 36.4\scalebox{0.8}{$\pm$0.5} \\
   & VAT\cite{miyato2018virtual} & TPAMI'19 &94.5\scalebox{0.8}{$\pm$0.3} & 90.4\scalebox{0.8}{$\pm$0.3} & 76.9\scalebox{0.8}{$\pm$0.6} & 75.0\scalebox{0.8}{$\pm$0.5} & 61.8\scalebox{0.8}{$\pm$0.4} & 59.6\scalebox{0.8}{$\pm$0.6} & 36.7\scalebox{0.8}{$\pm$0.5} & 36.3\scalebox{0.8}{$\pm$0.6} \\
   & FixMatch~\cite{sohn2020fixmatch} & NeurIPS'20 & - & - & 81.5\scalebox{0.8}{$\pm$0.2} & 80.9\scalebox{0.8}{$\pm$0.3} & 65.9\scalebox{0.8}{$\pm$0.3} & 65.2\scalebox{0.8}{$\pm$0.3} & - & - \\

   \midrule
   
   \multirow{10}{*}{\rotatebox{90}{OSSL}} & DS3L~\cite{guo2020safe} & ICML'20& 96.8\scalebox{0.8}{$\pm$0.3} & 94.5\scalebox{0.8}{$\pm$0.4} & 78.1\scalebox{0.8}{$\pm$0.4} & 76.9\scalebox{0.8}{$\pm$0.5} & - & - & - & - \\
   & UASD~\cite{chen2020semi} & AAAI'20 & 96.2\scalebox{0.8}{$\pm$0.6} & 94.3\scalebox{0.8}{$\pm$0.8} & 77.6\scalebox{0.8}{$\pm$0.4} & 76.0\scalebox{0.8}{$\pm$0.4} & 61.8\scalebox{0.8}{$\pm$0.4} & 58.4\scalebox{0.8}{$\pm$0.5} & 37.1\scalebox{0.8}{$\pm$0.7} & 36.9\scalebox{0.8}{$\pm$0.6} \\
   & MTCF~\cite{yu2020multi} & ECCV'20 & 93.7\scalebox{0.8}{$\pm$0.5} & 88.5\scalebox{0.8}{$\pm$0.3} & 85.5\scalebox{0.8}{$\pm$0.6} & 81.7\scalebox{0.8}{$\pm$0.5} & 63.1\scalebox{0.8}{$\pm$0.6} & 61.1\scalebox{0.8}{$\pm$0.3} & 37.0\scalebox{0.8}{$\pm$0.5} & 36.6\scalebox{0.8}{$\pm$0.4} \\
   & CL~\cite{cascante2021curriculum} & AAAI'21 & 96.9\scalebox{0.8}{$\pm$0.1} & 95.6\scalebox{0.8}{$\pm$0.4} & 83.2\scalebox{0.8}{$\pm$0.4} & 82.1\scalebox{0.8}{$\pm$0.4} & 63.6\scalebox{0.8}{$\pm$0.4} & 61.5\scalebox{0.8}{$\pm$0.5} & 37.3\scalebox{0.8}{$\pm$0.7} & 36.7\scalebox{0.8}{$\pm$0.8} \\
   & OpenMatch~\cite{saito2021openmatch} & NeurIPS'21 & 97.8\scalebox{0.8}{$\pm$0.2} & 96.0\scalebox{0.8}{$\pm$0.2} & 88.2\scalebox{0.8}{$\pm$0.2} & 85.5\scalebox{0.8}{$\pm$0.3} & 68.7\scalebox{0.8}{$\pm$0.1} & 68.4\scalebox{0.8}{$\pm$0.2} & 37.9\scalebox{0.8}{$\pm$0.4} & 37.0\scalebox{0.8}{$\pm$0.3} \\
   & T2T~\cite{huang2021trash} & ICCV'21 & 98.4\scalebox{0.8}{$\pm$0.1} & 96.2\scalebox{0.8}{$\pm$0.2} & 89.0\scalebox{0.8}{$\pm$0.4} & 86.9\scalebox{0.8}{$\pm$0.2} & 69.8\scalebox{0.8}{$\pm$0.2} & 68.0\scalebox{0.8}{$\pm$0.2} & 39.1\scalebox{0.8}{$\pm$0.3} & 37.3\scalebox{0.8}{$\pm$0.3} \\
   & SAFE-STUDENT~\cite{he2022safe} & CVPR'22& 98.3\scalebox{0.8}{$\pm$0.3} & 96.5\scalebox{0.8}{$\pm$0.1} & 85.7\scalebox{0.8}{$\pm$0.3} & 83.8\scalebox{0.8}{$\pm$0.1} & 68.4\scalebox{0.8}{$\pm$0.2} & 68.2\scalebox{0.8}{$\pm$0.1} & 37.7\scalebox{0.8}{$\pm$0.3} & 37.1\scalebox{0.8}{$\pm$0.3} \\
   & Ma~\etal~\cite{ma2023rethinking} & ICCV'23& \underline{98.7\scalebox{0.8}{$\pm$0.2}} & \underline{96.9\scalebox{0.8}{$\pm$0.1}} & \underline{91.4\scalebox{0.8}{$\pm$0.3}} & \underline{89.1\scalebox{0.8}{$\pm$0.1}} & \underline{72.5\scalebox{0.8}{$\pm$0.2}} & \underline{70.4\scalebox{0.8}{$\pm$0.1}} & \underline{40.8\scalebox{0.8}{$\pm$0.3}} & \underline{39.9\scalebox{0.8}{$\pm$0.3}} \\

   \midrule

   \rowcolor{green!30} & \textbf{SCOMatch} & \textbf{Ours} & \textbf{99.0\scalebox{0.8}{$\pm$0.1}} & \textbf{99.0\scalebox{0.8}{$\pm$0.1}} & \textbf{92.2\scalebox{0.8}{$\pm$0.2}} & \textbf{90.2\scalebox{0.8}{$\pm$0.6}} & \textbf{75.3\scalebox{0.8}{$\pm$0.5}} & \textbf{73.5\scalebox{0.8}{$\pm$0.2}} & \textbf{54.2\scalebox{0.8}{$\pm$0.3}} & \textbf{52.3\scalebox{0.8}{$\pm$0.3}} \\
   \bottomrule
\end{tabular}
}
\end{center}
\end{table*}

%% file: tabs/tab2.tex
\begin{table*}[t]
\centering
\begin{center}
\caption{AUC for OOD class identification of different methods on MNIST.}
\label{tab:tab2}
\resizebox{\linewidth}{!}{
\begin{tabular}{cc|cccccccc}
   \toprule
    \multicolumn{2}{c|}{Mismatch ratio} & 0.1 & 0.2 & 0.3 & 0.4 & 0.5 & 0.6 & Avg \\
   \midrule
   Probabilities~\cite{hendrycks2016baseline} & ICLR'17 &  84.3\scalebox{0.8}{$\pm$0.9} & 84.3\scalebox{0.8}{$\pm$0.9} & 84.3\scalebox{0.8}{$\pm$0.9} & 84.3\scalebox{0.8}{$\pm$0.9} & 84.3\scalebox{0.8}{$\pm$0.9} & 84.3\scalebox{0.8}{$\pm$0.9} & 84.3 \\
   DS3L~\cite{guo2020safe} & ICML'20 &  93.1\scalebox{0.8}{$\pm$0.4} & 91.7\scalebox{0.8}{$\pm$0.2} & 90.6\scalebox{0.8}{$\pm$0.1} & 90.5\scalebox{0.8}{$\pm$0.5} & 89.1\scalebox{0.8}{$\pm$0.2} & 85.1\scalebox{0.8}{$\pm$0.8} & 90.9 \\
   SAFE-STUDENT~\cite{he2022safe} & CVPR'22  &  97.3\scalebox{0.8}{$\pm$0.2} & 96.5\scalebox{0.8}{$\pm$0.1} & 96.0\scalebox{0.8}{$\pm$0.9} & 94.6\scalebox{0.8}{$\pm$0.9} & 93.5\scalebox{0.8}{$\pm$0.3} & 91.4\scalebox{0.8}{$\pm$0.2} & 95.3 \\
   Ma~\etal~\cite{ma2023rethinking}  & ICCV'23 & \textbf{98.0\scalebox{0.8}{$\pm$0.2}} & \underline{97.5\scalebox{0.8}{$\pm$0.1}} & \underline{97.0\scalebox{0.8}{$\pm$0.2}} & \underline{96.1\scalebox{0.8}{$\pm$0.4}} & \underline{94.8\scalebox{0.8}{$\pm$0.2}} & \underline{93.0\scalebox{0.8}{$\pm$0.2}} & \underline{96.4} \\
   \midrule
   \rowcolor{green!30} \textbf{SCOMatch} & \textbf{Ours} & \textbf{98.0\scalebox{0.8}{$\pm$0.2}} & \textbf{98.3\scalebox{0.8}{$\pm$0.2}} & \textbf{98.0\scalebox{0.8}{$\pm$0.1}} & \textbf{98.3\scalebox{0.8}{$\pm$0.2}} & \textbf{98.0\scalebox{0.8}{$\pm$0.1}} & \textbf{98.2\scalebox{0.8}{$\pm$0.2}} & \textbf{98.1} \\
   
   \bottomrule
\end{tabular}
}
\end{center}
\end{table*}

%% file: tabs/tab3.tex
\begin{table*}[t]
\centering
\begin{center}
\caption{Close-set classification accuracy (\%) of different methods.}
\label{tab:tab3}
\resizebox{\linewidth}{!}{
\begin{tabular}{cc|cc|cc|cc|c}
\toprule
\multicolumn{2}{c|}{Dataset} & \multicolumn{2}{c|}{CIFAR-10} & \multicolumn{4}{c|}{CIFAR-100} & IN-30\\
\midrule
\multicolumn{2}{c|}{No. of ID / OOD classes} & \multicolumn{2}{c|}{6 / 4} & \multicolumn{2}{c|}{55 / 45} & \multicolumn{2}{c|}{80 / 20} & 20 / 10  \\
\midrule
\multicolumn{2}{c|}{No. of labeled samples} & 25 & 50 & 25 & 50 & 25 & 50 & 10\%\\
\midrule
FixMatch~\cite{sohn2020fixmatch} & NeurIPS'20 & 91.9\scalebox{0.8}{$\pm$0.2} & 91.3\scalebox{0.8}{$\pm$0.2} & 69.9\scalebox{0.8}{$\pm$0.0} & 73.3\scalebox{0.8}{$\pm$0.6} & 63.6\scalebox{0.8}{$\pm$0.4} & 67.1\scalebox{0.8}{$\pm$0.1} & 87.1\scalebox{0.8}{$\pm$0.4} \\
MTCF~\cite{yu2020multi} & ECCV'20 & 71.9\scalebox{0.8}{$\pm$10.1} & 81.0\scalebox{0.8}{$\pm$5.2} & 58.1\scalebox{0.8}{$\pm$2.1} & 66.0\scalebox{0.8}{$\pm$0.8} & 52.3\scalebox{0.8}{$\pm$0.1} & 59.2\scalebox{0.8}{$\pm$0.0} & 82.4\scalebox{0.8}{$\pm$0.7}  \\
OpenMatch~\cite{saito2021openmatch} & NeurIPS'21 & 54.9\scalebox{0.8}{$\pm$2.3} & 91.3\scalebox{0.8}{$\pm$1.2} & 67.1\scalebox{0.8}{$\pm$1.4} & 71.9\scalebox{0.8}{$\pm$1.1} & 52.1\scalebox{0.8}{$\pm$4.8} & 66.9\scalebox{0.8}{$\pm$0.2} &
\underline{89.6\scalebox{0.8}{$\pm$1.0}}\\
T2T~\cite{huang2021trash} & ICCV'21 & 83.2\scalebox{0.8}{$\pm$1.0} & 90.6\scalebox{0.8}{$\pm$0.1} & 65.7\scalebox{0.8}{$\pm$0.9} & 70.7\scalebox{0.8}{$\pm$0.1} & 47.6\scalebox{0.8}{$\pm$10.4} & 64.2\scalebox{0.8}{$\pm$0.6} &
88.9\scalebox{0.8}{$\pm$0.1}\\
SSB~\cite{fan2023ssb} & ICCV'23 & 91.7\scalebox{0.8}{$\pm$0.2} & 92.2\scalebox{0.8}{$\pm$0.3} & \underline{70.6\scalebox{0.8}{$\pm$0.4}} & \underline{73.7\scalebox{0.8}{$\pm$0.8}} & \underline{64.2\scalebox{0.8}{$\pm$0.4}} & \underline{68.0\scalebox{0.8}{$\pm$0.2}} & - \\
IOMatch~\cite{li2023iomatch} & ICCV'23 & \underline{93.0\scalebox{0.8}{$\pm$0.1}} & \underline{93.3\scalebox{0.8}{$\pm$0.1}} & 69.2\scalebox{0.8}{$\pm$0.3} & 72.6\scalebox{0.8}{$\pm$0.3} & 63.7\scalebox{0.8}{$\pm$0.5} & 67.6\scalebox{0.8}{$\pm$0.1} & 89.2\scalebox{0.8}{$\pm$0.3} \\
\midrule
\rowcolor{green!30} \textbf{SCOMatch} & \textbf{Ours} & \textbf{94.1\scalebox{0.8}{$\pm$0.4}} & \textbf{94.0\scalebox{0.8}{$\pm$0.1}} & \textbf{71.1\scalebox{0.8}{$\pm$1.2}} & \textbf{74.3\scalebox{0.8}{$\pm$0.8}} & \textbf{64.4\scalebox{0.8}{$\pm$0.4}} & \textbf{70.0\scalebox{0.8}{$\pm$0.4}} &
\textbf{91.0\scalebox{0.8}{$\pm$0.1}} 
\\
\bottomrule
\end{tabular}
}
\end{center}
\end{table*}

%% file: tabs/tab4.tex
\begin{table*}[t]
\centering
\begin{center}
\caption{AUC for OOD class identification of different methods.}
\label{tab:tab4}
\resizebox{\linewidth}{!}{
\begin{tabular}{cc|cc|cc|cc|c}
\toprule
\multicolumn{2}{c|}{Dataset} & \multicolumn{2}{c|}{CIFAR-10} & \multicolumn{4}{c|}{CIFAR-100} & IN-30 \\
\midrule
\multicolumn{2}{c|}{No. of ID / OOD classes} & \multicolumn{2}{c|}{6 / 4} & \multicolumn{2}{c|}{55 / 45} & \multicolumn{2}{c|}{80 / 20} & 20 / 10 \\
\midrule
\multicolumn{2}{c|}{No. of labeled samples} & 25 & 50 & 25 & 50 & 25 & 50 & 10\% \\
\midrule
FixMatch~\cite{sohn2020fixmatch} & NeurIPS'20 & 37.4\scalebox{0.8}{$\pm$0.8} & 39.4\scalebox{0.8}{$\pm$0.2} & 54.5\scalebox{0.8}{$\pm$1.0} & 55.8\scalebox{0.8}{$\pm$1.0} & 41.4\scalebox{0.8}{$\pm$0.1} & 44.3\scalebox{0.8}{$\pm$0.8} & 88.6\scalebox{0.8}{$\pm$0.5} \\
MTCF~\cite{yu2020multi} & ECCV'20 & 92.0\scalebox{0.8}{$\pm$3.5} & 94.5\scalebox{0.8}{$\pm$2.0} & 76.9\scalebox{0.8}{$\pm$1.5} & 72.5\scalebox{0.8}{$\pm$0.2} & 69.2\scalebox{0.8}{$\pm$0.9} & 72.4\scalebox{0.8}{$\pm$1.9} & 93.8\scalebox{0.8}{$\pm$0.8}  \\
OpenMatch~\cite{saito2021openmatch} & NeurIPS'21 & 62.5\scalebox{0.8}{$\pm$4.2} & 99.4\scalebox{0.8}{$\pm$0.2} & 84.9\scalebox{0.8}{$\pm$0.1} & 87.0\scalebox{0.8}{$\pm$0.2} & 74.9\scalebox{0.8}{$\pm$3.8} & \textbf{86.2\scalebox{0.8}{$\pm$0.5}} &
\underline{96.4\scalebox{0.8}{$\pm$0.7}}\\
T2T~\cite{huang2021trash} & ICCV'21 & 34.9\scalebox{0.8}{$\pm$27.5} & 23.9\scalebox{0.8}{$\pm$8.5} & 53.0\scalebox{0.8}{$\pm$6.2} & 59.5\scalebox{0.8}{$\pm$1.5} & 50.5\scalebox{0.8}{$\pm$9.2} & 61.4\scalebox{0.8}{$\pm$21.1} &
84.5\scalebox{0.8}{$\pm$0.1} \\
SSB~\cite{fan2023ssb} & ICCV'23 &  \underline{99.4\scalebox{0.8}{$\pm$0.4}} & \underline{99.6\scalebox{0.8}{$\pm$0.2}} & \underline{89.4\scalebox{0.8}{$\pm$0.4}} & \underline{90.6\scalebox{0.8}{$\pm$0.5}} & \underline{90.3\scalebox{0.8}{$\pm$1.3}} & 85.3\scalebox{0.8}{$\pm$2.1}  & - \\
IOMatch~\cite{li2023iomatch} & ICCV'23 &  53.5\scalebox{0.8}{$\pm$0.1} & 60.6\scalebox{0.8}{$\pm$1.0} & 70.2\scalebox{0.8}{$\pm$0.3} & 71.9\scalebox{0.8}{$\pm$0.4} & 63.2\scalebox{0.8}{$\pm$0.2} & 63.9\scalebox{0.8}{$\pm$1.7}  & 89.4\scalebox{0.8}{$\pm$0.5}\\
\midrule
\rowcolor{green!30} \textbf{SCOMatch} & \textbf{Ours} & \textbf{99.9\scalebox{0.8}{$\pm$0.0}} & \textbf{99.7\scalebox{0.8}{$\pm$0.3}} & \textbf{90.3\scalebox{0.8}{$\pm$0.8}} & \textbf{91.3\scalebox{0.8}{$\pm$0.4}} & \textbf{94.0\scalebox{0.8}{$\pm$1.6}} & \underline{85.4\scalebox{0.8}{$\pm$0.3}} & 
\textbf{97.5\scalebox{0.8}{$\pm$0.5}} \\
\bottomrule
\end{tabular}
}
\end{center}
\end{table*}

%% file: tabs/tab6.tex
\begin{table*}[t]
\begin{center}
\caption{Ablation studies of the head structure and the SCO self-training in SCOMatch. The model is trained on CIFAR-10 (6 ID classes) and CIFAR-100 (55 ID classes) with 25 labeled images per class. Column 1 indicates FixMatch.}
\label{tab:tab6}
\begin{tabular}{c|cc|ccc|ccc}
\toprule
 \multirow{2}{*}[-0.8ex]{Single head}& \multicolumn{2}{c|}{SCO self-training} & \multicolumn{3}{c|}{CIFAR-10} & \multicolumn{3}{c}{CIFAR-100} \\
\cmidrule{2-9}
& $\mathcal{L}^{close}_{u}$ & $\mathcal{L}^{open}_{u}$ & ACC$_{close}$ & ACC$_{open}$ & AUC & ACC$_{close}$ & ACC$_{open}$ & AUC \\
\midrule
 &           &            & 
 91.7 & 55.0 & 37.1 & 69.9 & 38.4 & 53.5 \\
 
             & \checkmark & \checkmark & 
 90.5 & 54.3 & 61.4 & 69.1 & 39.7 & 72.1 \\
 
  \checkmark &            & \checkmark & 
 93.2 & 95.2 & 99.0 & 68.7 & 71.6 & 88.5 \\
  \checkmark &  \checkmark &            & 
 93.5 & 80.6 & 90.9 & 69.0 & 58.7 & 75.0 \\
\midrule
  \checkmark &  \checkmark & \checkmark & 
 \textbf{94.4} & \textbf{95.4} & \textbf{99.3} & \textbf{73.3} & \textbf{74.4} & \textbf{90.6} \\
\bottomrule
\end{tabular}
\end{center}
\end{table*}